\def\year{2020}\relax
\documentclass[letterpaper]{article} 
\usepackage{aaai20}  
\usepackage{times}  
\usepackage{helvet} 
\usepackage{courier}  
\usepackage[hyphens]{url}  
\usepackage{graphicx} 
\usepackage{graphicx}  
\usepackage{booktabs}       
\usepackage{varwidth}       
\usepackage{bm} 

\title{Semantics to Space(S2S): Embedding Semantics Into Spatial Space For Zero-shot Verb-object Query Inferencing }
\author{
\Large \textbf{Sungmin Eum\textsuperscript{\rm 1,2}, Heesung Kwon\textsuperscript{\rm 1}}\\
\textsuperscript{\rm 1}U.S. Army Research Lab\\
\textsuperscript{\rm 2}Booz Allen Hamilton\\
 \\ 
}
\begin{document}

\maketitle

\begin{abstract}
We present a novel deep zero-shot learning (ZSL) model for inferencing human-object-interaction with verb-object (VO) query. While the previous two-stream ZSL approaches only use the semantic/textual information to be fed into the query stream, we seek to incorporate and embed the semantics into the visual representation stream as well. Our approach is powered by Semantics-to-Space (S2S) architecture where semantics derived from the residing objects are embedded into a spatial space of the visual stream. This architecture allows the co-capturing of the semantic attributes of the human and the objects along with their location/size/silhouette information. To validate, we have constructed a new dataset, Verb-Transferability 60 (VT60). VT60 provides 60 different VO pairs with overlapping verbs tailored for testing two-stream ZSL approaches with VO query. Experimental evaluations show that our approach not only outperforms the state-of-the-art, but also shows the capability of consistently improving performance regardless of which ZSL baseline architecture is used.
\end{abstract}

\section{Introduction}

While computer vision has shown great successes in areas, such as object detection \cite{Ren2015Faster,li2017light,Redmon2018yolo} or recognition\cite{He16Deep,Simonyan15Very}, visual scene understanding is still a very challenging area. Image caption generation is one of the early attempts in providing scene description of an image. 
Although some good results have been reported in \cite{Yao2017BoostingIC,Vinyals2015Show}, the depth of explanation is shallow since they are often anchored on statistical correlations rather than semantic relationships. What is required here is of a method rooted more closely on semantics of entities in the scene and their relationships. Compared to other sensory modality, vision provides a distinctive advantage of providing spatial information of entities captured in the field of view. 

Our assertion is that better semantic understanding of a scene can be obtained by exploiting spatial relationships among the entities and their semantics recognized in the image. This, however, poses another challenge in terms of creating sufficiently large labeled dataset for training. Since there can be a very large variety of objects that can be present in a scene (depending on the task), a combinatorial explosion of different relationships among the objects can also be expected. Such sparse dataset problem is a persistent issue in supervised learning, and there have been a series of recent approaches where few-shot/zero-shot \cite{sung2018learning,ba2015Predicting,Socher2013Zero,Vinyals2016Matching,Ravi2017Optimization,Romera-paredes2015An,Zhang2016ZeroShotLV,Lampert2014AttributeBasedCF} learning methods have been explored.

In short, in order to better tackle semantic understanding of different sparsely occurring human-involved scenes (even the unseen cases), we observe the necessity to focus on two major aspects: being able to make use of relative {\it spatial relationships among the semantics}, and being able to robustly handle hardly encountered or unseen scenes. However, none of the previous ZSL approaches seek to analyze and exploit spatial relationship {\it directly} across the semantics. Instead, they commonly ground their approaches on a two-stream network where RGB image is used as visual input while semantics (textual descriptions or attributes) are used for query. This motivated us to consider injecting semantic information into both of the streams. Note that the end goal is to be able to accurately tell whether the visual input and the query are a good match or not. Therefore, injecting the semantic information on both sides can be interpreted as bringing the starting points closer to each other before we learn the {\it closeness score} between the two.

\begin{figure}[h]
\centering
\includegraphics[width=\linewidth]{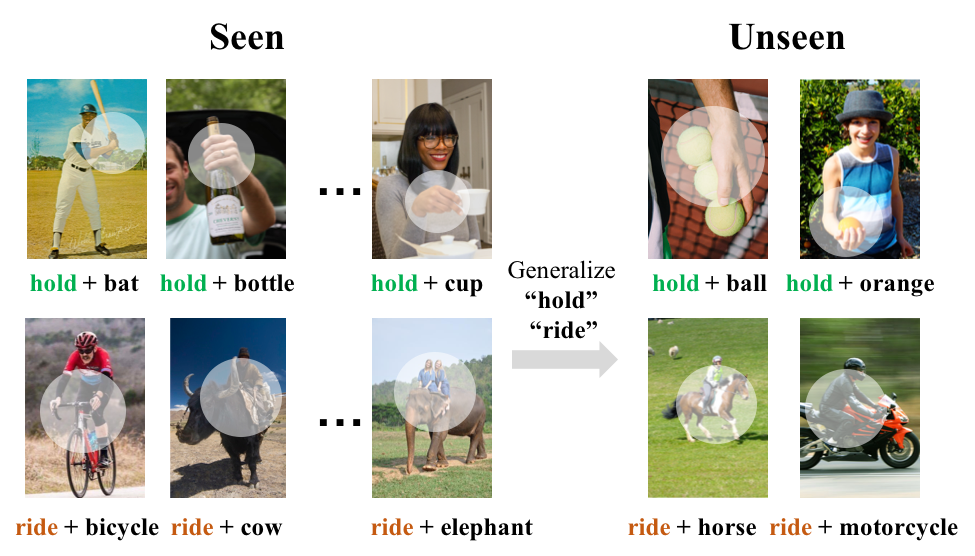}
\caption{{\bf Zero-shot VO Inferencing.} From the seen dataset, human-involved action verbs such as `hold' or `ride' are generalized so that the trained model can be used for zero-shot verb-object query inferencing on unseen Verb-Object (VO) pairs with the same verbs and different objects.}
\label{fig:zeroshotInference}
\end{figure}

Accordingly, we introduce a novel architecture that enables the direct embedding of semantic information into local regions in an image occupied by salient foreground objects in such a way that both the semantics and the spatial attributes of the objects are jointly embedded into the visual input stream.  The co-embedding of the semantics and spatial attributes of objects, referred to as Semantic-to-Space (S2S) embedding, can induce much stronger correlations between the newly engineered visual input and the corresponding semantic query and thus is found to be central to enhancing overall performance of the whole network targeted to perform zero-shot Verb-Object (VO) query (See Figure \ref{fig:zeroshotInference}).

One of the main strengths of the proposed work is that the S2S embedding is highly suitable for encoding the semantics from the very early stage of the CNN pipeline, such as implicit actions and/or states, associated with the spatial relationships and interactions among multiple objects co-occurring in the scene. This can also encode various geometrical properties of foreground entities in terms of size, degrees of occlusions, object silhouettes, etc., making the joint embedding quite powerful in representing the scene.

To validate the effectiveness of our approach, we have constructed a new dataset, Verb-Transferability 60 (VT60). VT60 contains human-involved scenes (Figure \ref{fig:VT60Dataset}) which provides 60 different VO pairs with overlapping verbs tailored for testing ZSL approaches with VO query. Through experimental evaluations, we show that our novel architecture not only can outperform the two-stream zero-shot state-of-the-arts in the VO query setting, but also capable of being able to generate an effective feature (S2S) which consistently improves upon representative baseline architectures.

\section{Relevant Work}
Among numerous zero-shot classification/recognition approaches that have been shared with the community over the past several years, we constrain our focus on two-stream zero-shot approaches \cite{Frome2013Devise,Akata2015Evaluation,Yang2015Unified,Zhang2015ZeroShotLV,zhang2017learning} which typically include a visual stream, a query stream, and a module that computes whether the features from the two streams (i.e., encoded visual input and the query) are an appropriate match.

After noting that the two-stream zero-shot approaches \cite{Frome2013Devise,Socher2013Zero,ba2015Predicting,Yang2015Unified,zhang2017learning,sung2018learning} were more attracted to figuring out how to manipulate the aggregation of the two features (or {\it to construct a joint embedding space}) at the posterior portion of the streams, our initial question which led the overall direction of this study was ``Why are the {\it semantics} only exploited for constructing query features but {\it not} for generating visual features when the end goal is to train a module which is supposed to learn to match the two?'', and ``Will the semantics be effective in better learning the joint embedding space if embedded into both the visual and the semantic stream?''

\begin{figure}[h]
\centering
\includegraphics[width=\linewidth]{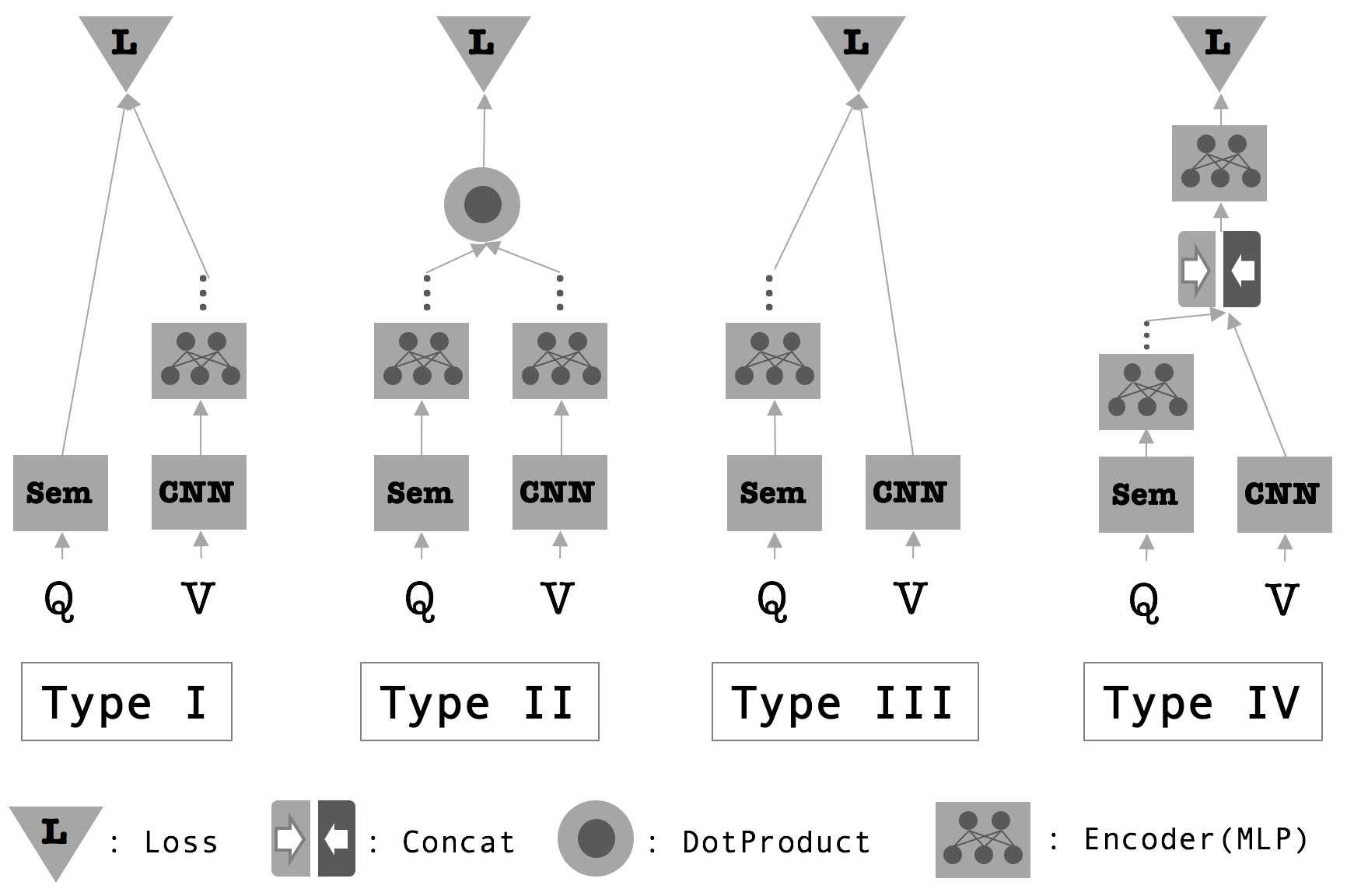}
\caption{{\bf Representative joint embedding types.} Illustrations of representative two-stream zero-shot learning architectures. Q and V are query and visual input, respectively. Sem and CNN indicate semantics encoder (e.g., W2V) and CNN-based encoder, respectively. Type I=\cite{Frome2013Devise,Socher2013Zero}, II=\cite{Yang2015Unified,ba2015Predicting}, III=\cite{zhang2017learning}, IV=\cite{sung2018learning}}
\label{fig:jointEmbeddingTypes}
\end{figure}

Accordingly, we constrain our scope and fixate more on devising a way to {\it inject the semantics into the visual stream} and validate the effectiveness of semantics-embedded visual feature (S2S), rather than to thoroughly compare our overall system with an exhaustive list of state-of-the-art zero-shot approaches. Figure \ref{fig:jointEmbeddingTypes} shows the symbolic illustrations of representative two-stream zero-shot learning approaches. In all the previous Types, visual features are constructed by feeding regular RGB images into conventional CNN encoders resorting to only visual information. Under our S2S scheme, in contrast, visual features are constructed by augmenting the visual information with spatial semantics.

\section{Approach}

\begin{figure*}[t]
\begin{center}
\includegraphics[width=0.60\linewidth]{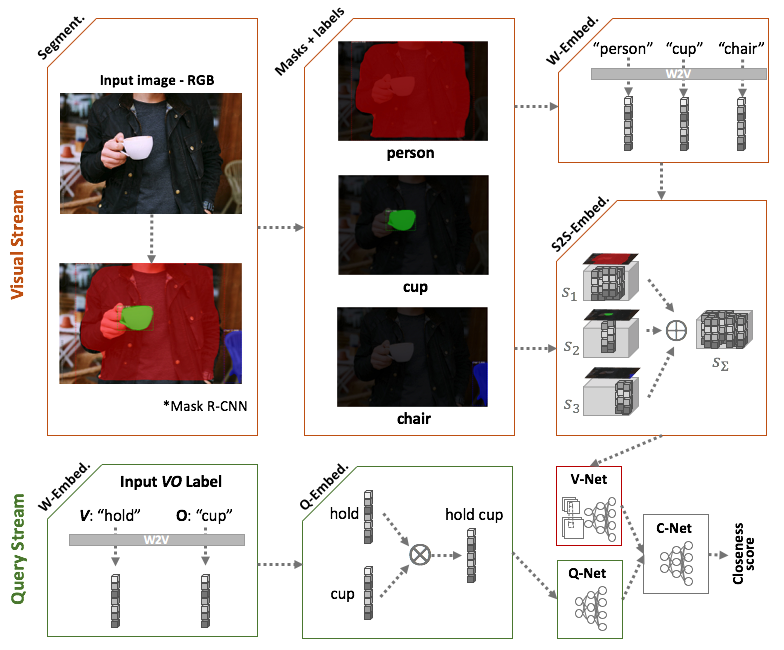}
\caption{{\bf Embedding the semantics into the spatial space.} Semantic segmentation-driven object masks are used to generate object-wise feature blobs ($s_{i}$) which are then aggregated to create one S2S feature blob ($s_{\Sigma}$) for a given image.}
\label{fig:S2S_detail}
\end{center}
\end{figure*}

\subsection{Embedding the semantics into spatial space}
Following the spirit of the previous approaches \cite{ba2015Predicting,zhang2017learning,sung2018learning} designed for zero-shot or few-shot learning, our architecture retains two separate input streams, one for the visual input and the other for the semantic query/class descriptions (e.g., word embeddings such as Word2Vec \cite{Mikolov2013Distributed} or GloVe \cite{pennington2014glove}). While various technical approaches already exist on how and where two different feature spaces (visual and semantics) are co-embedded, we build our architecture to be similar to a model \cite{sung2018learning} where a transferrable deep metric for {\it computing the closeness} between the features originating from two different spaces is learned. 

In our approach, we do not jump into the competition of searching for a subspace which could accommodate better grounds for co-learning RGB and semantic features. Instead, we directly embed the semantics into the spatial space of the visual stream according to the auxiliary object information. Figure \ref{fig:S2S_detail} illustrates the overall architecture for our S2S approach. 

\noindent{{\bf Visual Stream.} For a given RGB image, semantic segmentation is carried out to acquire pixel-level masks and their semantic labels for the detected objects. We have utilized Mask R-CNN \cite{He2017MaskR} which provides reliable segmentation results. However, as long as the segmentation provides masks and semantic labels, any semantic segmentation can be used. Semantic `textual' labels are then fed into a word embedder (denoted as {\it W-Embed}) which computes the corresponding word representation vectors ($v_{i} \in \mathcal{R}^{l_{v}}$ ). Lastly, in {\it S2S-Embed}, each word representation vector is used to fill all the pixel locations within the corresponding object mask, generating a three-dimensional object-wise feature blob ($s_{i}$) with the size of $H\times W\times l_{v}$. Note that the vectors are placed so that they are perpendicular to the 2-D plane ($H \times W$) as depicted in Figure \ref{fig:S2S_detail}. All the object-wise feature blobs are summed along the depth axis creating one single S2S feature blob $s_{\Sigma}$. Thus, this blob is basically a set of semantic features embedded into local regions occupied by objects co-representing semantics and object silhouettes.

The process of adding the object-wise feature blobs can be seen as adding the word representation vectors at each pixel location if more than one objects are co-occurring. This is one unique advantage of employing S2S embedding as overlapping objects or object parts can be encoded in the form of added word representations. The act of adding word representation vectors generates a new, yet meaningful set of vectors as have been shown by Mikolov et al.  \cite{Mikolov2013Distributed} with Word2Vec vector addition. This eventually provides essential features throughout the spatial space which is found to be able to represent internal actions or states associated with multiple objects co-occurring in an image. The novel representation of implicit actions or states depicted in the image plays a key role in pulling up the zero-shot VO query inferencing.

A single S2S feature blob $s_{\Sigma}$ is fed into the Visual-Net (${\it V-Net}$) $f_{V}$, where the feature blob is being encoded as an effective deep feature $f_{V}(s_{\Sigma})$. We have used one of the models in the ResNet \cite{He16Deep} family as a visual encoder. Evaluations on different models will be presented in the Experiments (Section \ref{sec:Experiments}). Since the S2S input feature blob has a different channel dimension ($l_{v}$) when compared with that of an RGB image, we have made an architectural modification in the first layer of the encoder to accommodate the change.
}

\noindent{{\bf Query Stream.} 
To generate a query, we consider a VO phrase which consists of a verb phrase (V for verb) and an object phrase (O for object). Subject is omitted in constructing our query as a person is assumed to be the subject in all of our target VO images. As have been employed for the {\it W-Embed}. in the Visual Stream, query words for V and O are first converted to word representation vectors ($v_{V}, v_{O} \in \mathcal{R}^{l_{v}}$ ). These vectors are then combined together as one single query vector $v_{V+O}$ (see {\it Q-Embed.} in Figure \ref{fig:S2S_detail}). Summation of the word representation vectors is one way of aggregating the VO information. We also test out several variations on how these V and O vectors are generated within {\it Q-Embed.} which will be presented in an ablation study in Section \ref{subsec:QNetGeneration}. $v_{V+O}$ is then fed into the Query-Net ({\it Q-Net}) $f_{Q}$ which consists of fully connected layers and performs the encoding of the feature, pruducing $f_{Q}(v_{V+O})$.
}

\noindent{{\bf Computing the Closeness.}
The concatenated version of {\it V-Net} and {\it Q-Net} output features, $\Delta(f_{V}(s_{\Sigma}),f_{Q}(v_{V+O}))$, are fed into the Closeness Network ({\it C-Net}) $f_{C}$ which is designed to compute how close or relevant the visual input and the query is. {\it C-Net} produces a scalar {\it closeness} value $\tau_{V,Q} \in \{ 0,1 \}$ as below:

\begin{equation}
\tau_{V,Q} = f_{C}(\Delta(f_{V}(s_{\Sigma}),f_{Q}(v_{V+O})))
\end{equation}

}

Mean squre error (MSE) loss is used to regress $\tau_{V,Q}$ to the ground truth which is set to be 1 or 0 for matched and unmatched pairs, respectively. 

\subsection{Implementation Details}
\label{ImplementDetails}

We used a pretrained Mask R-CNN \cite{He2017MaskR} model, trained on MS COCO Dataset \cite{MSCOCO}, to acquire the object masks and their semantic labels. The model can detect 80 different object categories. For generating the word description vectors used in {\it W-Embed.} and {\it Q-Embed.} (Figure \ref{fig:S2S_detail}), we used a pretrained Word2Vec model \cite{W2V} trained on Google News dataset. This model provides 300-dimensional vectors for 3 million words and phrases.

We have inherited the baseline architectural framework of a ZSL approach by Sung et al. \cite{sung2018learning}. For generating the encoded vectors in {\it V-Net}, a ResNet \cite{He16Deep} (pretrained on ImageNet) model was employed. Since S2S feature blobs ($s_{\Sigma}$) differ in the number of channels when compared to conventional RGB images, we have modified the first layer of ResNet so as to fit the channel size of S2S and carried out a fine-tuning on the train set of VT60. We take the top pooling units of ResNet as the output of {\it V-Net}.

{\it Q-Net} was implemented with a Multi-layer Perceptron (MLP) where the input size is 300, matching the dimension of the Word2Vec features we use. The output dimension of {\it Q-Net} was set to match the output dimension of {\it V-Net} (512 for ResNet18 and 34, 2048 for ResNet50). {\it C-Net} is also implemented with MLP where the size of the hidden layer is set to be 1024 or 4096 for ResNet18 and ResNet 50, respectively. For the last layer of {\it C-Net}, we use Sigmoid to generate the {\it closeness score} which provides how close the visual input and the semantic query is.

Similar to \cite{sung2018learning}, we use weight decay (L2 regularization) in the fully connected layers of {\it Q-Net}, but not in the {\it R-Net}. Our S2S model along with all the existing models are trained with weight decay $10^{-5}$. We use the initial learning rate of $10^{-5}$ with Adam \cite{Kingma2015AdamAM} and annealed by 0.5 for every 200k iterations. Batch size for training is set to be 32.

We have employed episode-based zero-shot training strategy \cite{sung2018learning} for S2S and Orthovec2S models in the experiments. All the experiments are implemented using PyTorch \cite{PyTorch}. Codes for our approach and the Dataset will be publicly available online.

{\noindent {\bf Orthovec2S.}} In order to validate that the {\it semantic} vectors embedded into the S2S function as an independent factor in enhancing the overall performance (apart from having a {\it spatial} aspect), we have devised a similar architecture called Orthovec2S, where we use a set of orthonormal vectors (which does {\it not} carry any semantic information) in place of Word2Vec vectors.

\begin{figure*}[hbt]
\begin{center}
\includegraphics[width=0.90\linewidth]{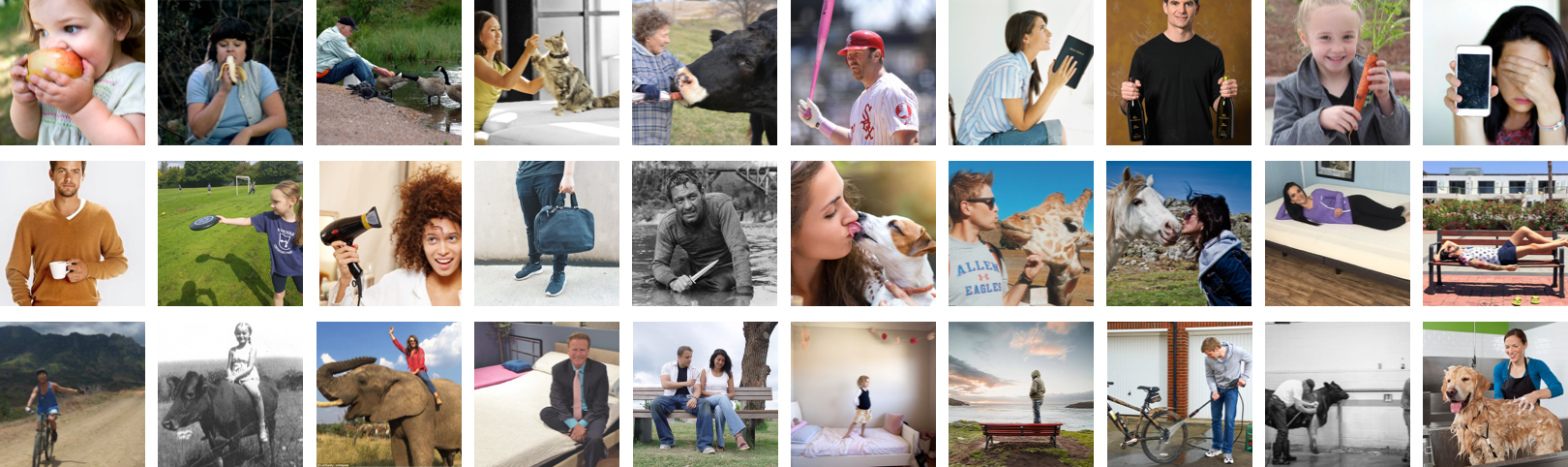}
\caption{{\bf VT60 Dataset.} One example image from each VO category of the Train Set is included. The images are arranged (top to bottom, left to right) to align with the category order provided in Table \ref{tab:VT60DatasetLabels}. For presentation purposes, each image is cropped with 1:1 aspect ratio.}
\label{fig:VT60Dataset}
\end{center}
\end{figure*}

\begin{table*}[h]
\caption{{\bf Train and Test set labels in VT60.} 60 VO pairs for Train (30) and Test (30) set.}
\begin{center}
\begin{tabular}{ ccc }
\toprule
Verbs & Train & Test \\
\midrule
{\bf eat}  & apple, banana  & broccoli, donut\\
{\bf feed}  & bird, cat, cow & dog, giraffe, horse, sheep \\
{\bf hold}  & baseball bat, book, bottle,  & orange, scissors, skateboard, \\ 
 & carrot, cell phone, cup, &  sports ball, surfboard, tennis racket,\\
 & frisbee, hair dryer, handbag, knife & toothbrush, vase, wine glass
\\
{\bf kiss}  & dog, giraffe, horse & bird, cat, cow\\
{\bf lie on}  & bed, bench  & couch, surfboard\\
{\bf ride}  & bicycle, cow, elephant & horse, motorcycle, sheep\\
{\bf sit on}  & bed, bench & chair, couch\\
{\bf stand on}  & bed, bench & chair, couch\\
{\bf wash}  & bicycle, cow, dog & elephant, horse, motorcycle\\
\bottomrule
\end{tabular}
\end{center}
\label{tab:VT60DatasetLabels}
\end{table*}

\section{Experiments}
\label{sec:Experiments}

\subsection{Dataset: Verb-Transferability 60}


None of the available dataset provides sufficient number of images and sufficient types of objects tied with each verb to train and test two-stream zero-shot HOI approaches where each image is preferred to have one semantic VO label. Accordingly, we have newly constructed a dataset, Verb-Transferability 60 (VT60) which contains images with 60 different human-involved verb-object (VO) pairs. As we seek to evaluate the capability of being able to identify images with unseen verbs (actions) + seen/unseen objects, multiple sets of VO images were collected where different verbs are paired with a same object (i.e., feed dog, feed giraffe, feed horse, feed sheep). In the dataset, nine different verbs (eat, feed, hold, lie on, stand on, etc.) and thirty seven objects (apple, bench, couch, elephant, and etc.) are involved to create 60 different VO pairs as listed in Table \ref{tab:VT60DatasetLabels}. Typical stationary verbs or verb phrases which are hard to be controversial when inspecting an image are included. We have collected 50 images for each VO pair, constructing 3000 images in the overall dataset. The images included in the dataset have been downloaded from the web via keyword-driven (e.g., ``riding horse'') image search. Example images from the train set of VO60 are shown in Figure \ref{fig:VT60Dataset}.

\begin{table}[ht]
    \centering
    \caption{{\bf Evaluating VO Confusion.} The train and test set do not share any verb-object combination while verbs and objects involved are exactly alike. Numbers indicate recognition accuracy in [\%].}
    \begin{tabular*}{\linewidth}{c@{\extracolsep{\fill}} c c }
      \toprule
      RGB & OrthoVec2S & S2S \\
      \midrule
      62.00 & 72.30 & {\bf 82.50} \\
      \bottomrule
    \end{tabular*}
    \label{table:VOConfusion}
\end{table}

\begin{figure}[h]
\begin{center}
\includegraphics[width=0.99\linewidth]{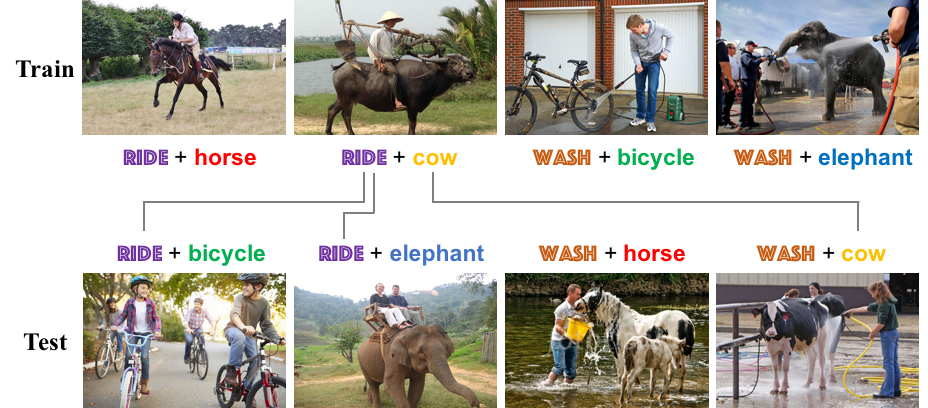}
\caption{{\bf VO Confusion Dataset.} Example images from each VO category of the VO Confusion Dataset are displayed.}
\label{fig:VOConfusionDataset}
\end{center}
\end{figure}

\subsection{Zero-shot I: Verb transferability}
\label{subsec:zeroshotI}
We train the network on the Train set of VT60 and evaluate on Test set. Both sets contain equivalent set of verbs but none of the VO pairs exist on both sides, which indicates that the evaluation would be focusing on how well the network learns to transfer the `seen' verbs paired with `unseen' objects (Thus, {\it Verb Transferability}). For instance, when testing an image of `Eat broccoli', nine different queries are generated by fixing the seen object while providing diversity in the action verbs. Among those queries, the one with the highest score gets the hit. The verb transferability performance comparison is shown in Table \ref{tab:verbTrans}. Regardless of which backbone network is used, S2S consistently outperforms other baselines.

\subsection{General Applicability of S2S}
\label{subsec:Effectiveness_S2S}
To validate the effectiveness and applicability of exploiting S2S which is not limited to the state-of-the-art architecture \cite{sung2018learning}, we have equipped other recently introduced ZSL approaches (see joint embedding Type I, II, III and IV in Figure \ref{fig:jointEmbeddingTypes}) with our S2S module for evaluation. Table \ref{fig:GeneralApplicability} shows that employing S2S embedding consistently improves zero-shot performance on VT60 dataset, validating the general applicability of our approach. Note that this also serves as a validation that our approach outperforms previously introduced representative approaches \cite{Socher2013Zero,ba2015Predicting,zhang2017learning,sung2018learning} including the state-of-the-art. Train/test setting used in Section \ref{subsec:zeroshotI} has equivalently been used and ResNet18 was used as the base architecture for all the cases. For this comparison, {\it Zero-shot I} scenario has been employed.

\begin{table}[htb]
\small
\caption{ {\bf General applicability of S2S.} Regardless of the baseline architecture used (all RGB input), S2S embedding consistently boosts the zero-shot accuracy. Baseline architectures are Socher et al. \cite{Socher2013Zero}, Ba et al. \cite{ba2015Predicting}, \cite{zhang2017learning}, Sung et al. \cite{sung2018learning} which are selected for the four joint embedding Types. All the numbers indicate recognition accuracy in [\%].}
\begin{center}
\begin{tabular*}{\linewidth}{c@{\extracolsep{\fill}}cccc}
\toprule
Type & Model & w/o S2S & w/ S2S & Gain \\
\midrule
I & Socher et al. & 27.13 & {\bf 30.00} & $+$2.87 \\
II & Ba et al. & 22.13 & {\bf 29.73} & $+$7.60 \\
III & Zhang et al. & 19.27 & {\bf 24.20} & $+$4.93 \\
IV & Sung et al. & 33.53 & {\bf 40.40} & $+$6.87 \\
\bottomrule
\end{tabular*}
\end{center}
\label{fig:GeneralApplicability}
\end{table}

\begin{table}[t]
\small
\caption{{\bf Verb transferability evaluation.} RGB: Sung et al. \cite{sung2018learning}, OrthoVec2S: Orthonormal Vectors-to-space, S2S: Semantics-to-space. All the numbers indicate recognition accuracy in [\%].} 
\begin{center}
\begin{tabular*}{\linewidth}{c@{\extracolsep{\fill}}ccc}
\toprule
Architecture & RGB & OrthoVec2S & S2S \\
\midrule
ResNet18  & 33.53 & 40.40 & {\bf 46.27} \\
ResNet34 & 38.00 & 43.53 & {\bf 48.87} \\
ResNet50  & 41.73 & 44.60 & {\bf 50.47} \\
\bottomrule
\end{tabular*}
\end{center}
\label{tab:verbTrans}
\end{table}

\subsection{Zero-shot II: Confusing verb-object pair scenario}
\label{subsec:zeroshotII}
In this scenario, we focus on a slightly different case where the verbs and the objects are confusingly paired. We have constructed a mini-dataset (VO Confusion Dataset shown in Figure \ref{fig:VOConfusionDataset}) where only two verbs (ride and wash) and four objects (horse, cow, bicycle and elephant) are involved. For each VO pair, we have collected 100 images which sums up to 800 images in overall. This setting is devised in order to evaluate if the network is putting more focus on the object than the verbs (i.e., actions) in the scene. For instance, when we train the network to recognize ``ride+horse'' and ``wash+bicycle'', will it be able to correctly recognize when ``ride+bicycle'' or ``wash+horse'' images are presented in testing? Table \ref{table:VOConfusion} shows that the baseline architecture \cite{sung2018learning} does carry slight capability (accuracy = 62$\%$) in avoiding the confusion. However, it is outperformed by both OrthoVec2S and S2S (82.50$\%$), indicating that the proposed embedding also helps in lessening the confusion in a very common zero-shot scenario.

\subsection{Discriminative power analysis}
We analyze the discriminative power of the features acquired from the models (RGB \cite{sung2018learning}, Orthovec2s, and S2S) involved in the comparison using tSNE \cite{vanDerMaaten2008} visualization. Features acquired at the end of either {\it V-Net} or {\it Q-Net} (see Figure \ref{fig:S2S_detail}) are considered for the analyses.

\begin{figure}[!htb]
\begin{center}
\includegraphics[width=\linewidth]{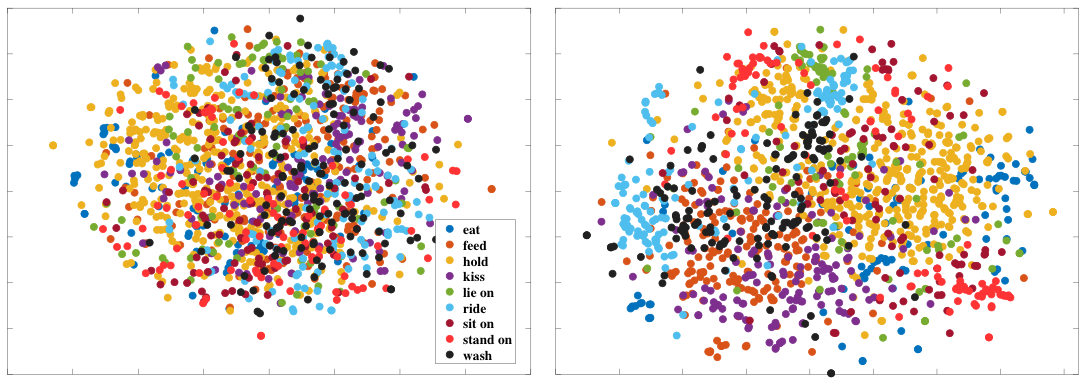}
\caption{{\bf V-Net Feature ($\bm{f_{V.Net}}$) Distribution Visualization (t-SNE): Verb-wise.} For each verb, the {\it S2S} (right) V-Net features are better clustered than those extracted from {\it RGB}\protect\cite{sung2018learning} (left) architecture.}
\label{fig:VNet_Feat_Comparison_Verbwise}
\end{center}
\end{figure}

\noindent{\bf V-Net Features: Verb-wise Analysis.} Figure \ref{fig:VNet_Feat_Comparison_Verbwise} shows that the {\it V-Net} features ($\bm{f_{V.Net}}$) from S2S architecture are more tightly aggregated and clustered with respect to each action verb when compared to the case of the RGB architecture \cite{sung2018learning}. Notice that in the RGB case, many different verb features are mingled together in the center region of the plot presenting that these features are less discriminative in terms of the actions (i.e., verbs) depicted in the images. As each $\bm{f_{V.Net}}$ gets concatenated with the corresponding $\bm{f_{Q.Net}}$ (i.e., query feature which is driven by action verbs) before it is fed into {\it R-Net}, disentangling the features according to the action verbs eventually provides better learning ground for the {\it R-Net}.

\begin{figure*}[t]
\begin{center}
\includegraphics[width=0.85\linewidth]{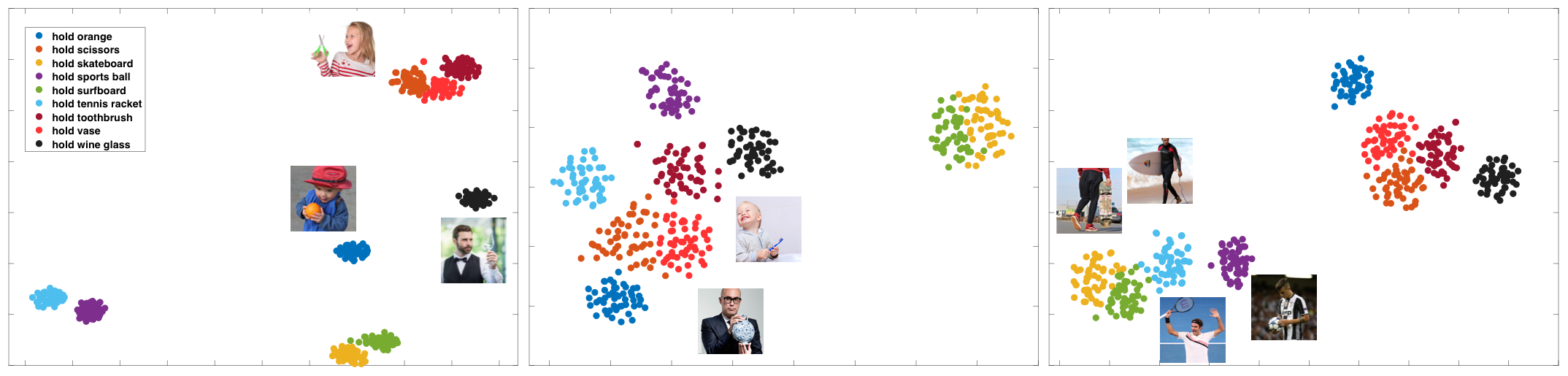}
\caption{{\bf $\bm{f_{V.Net}\oplus f_{Q.Net}}$ Feature Distribution Visualization (t-SNE): Class-wise.} Concatenated features of V-Net and Q-Net outputs acquired from {\it RGB}\protect\cite{sung2018learning} (left), {\it Orthovec2S} (center), and {\it S2S} (right) models. Only the samples corresponding to ``hold''-related classes from the Test Set are plotted.}
\label{fig:VNetQNet_Feat_Comparison_Classwise}
\end{center}
\end{figure*}

\noindent{\bf Q-Net + V-Net: Class-wise Analysis.}
In Figure \ref{fig:VNetQNet_Feat_Comparison_Classwise}, Concatenated features ($\bm{f_{V.Net}}$ and $\bm{f_{Q.Net}}$) for VO pairs that is tied with the verb ``hold'' are depicted for all three architectures. In the S2S case, we can observe that sports-related VO pairs, i.e., ``hold sports ball'', ``hold tennis racket'', ``hold surfboard'', ``hold skateboard'' features are located in close proximity showing that the semantics are properly embedded along with the visual attributes. In contrast, under the RGB architecture \cite{sung2018learning}, ``hold skateboard/surfboard'' is located closer to ``hold orange'' than other sports-related VO pairs. Comparing S2S with orthovec2S indicates that embedding actual semantics into the visual stream (instead of synthesized orthogonal vectors) helps more in learning the features which better reflect the underlying semantics.

\begin{figure*}[!htb]
\begin{center}
\includegraphics[width=0.90\linewidth]{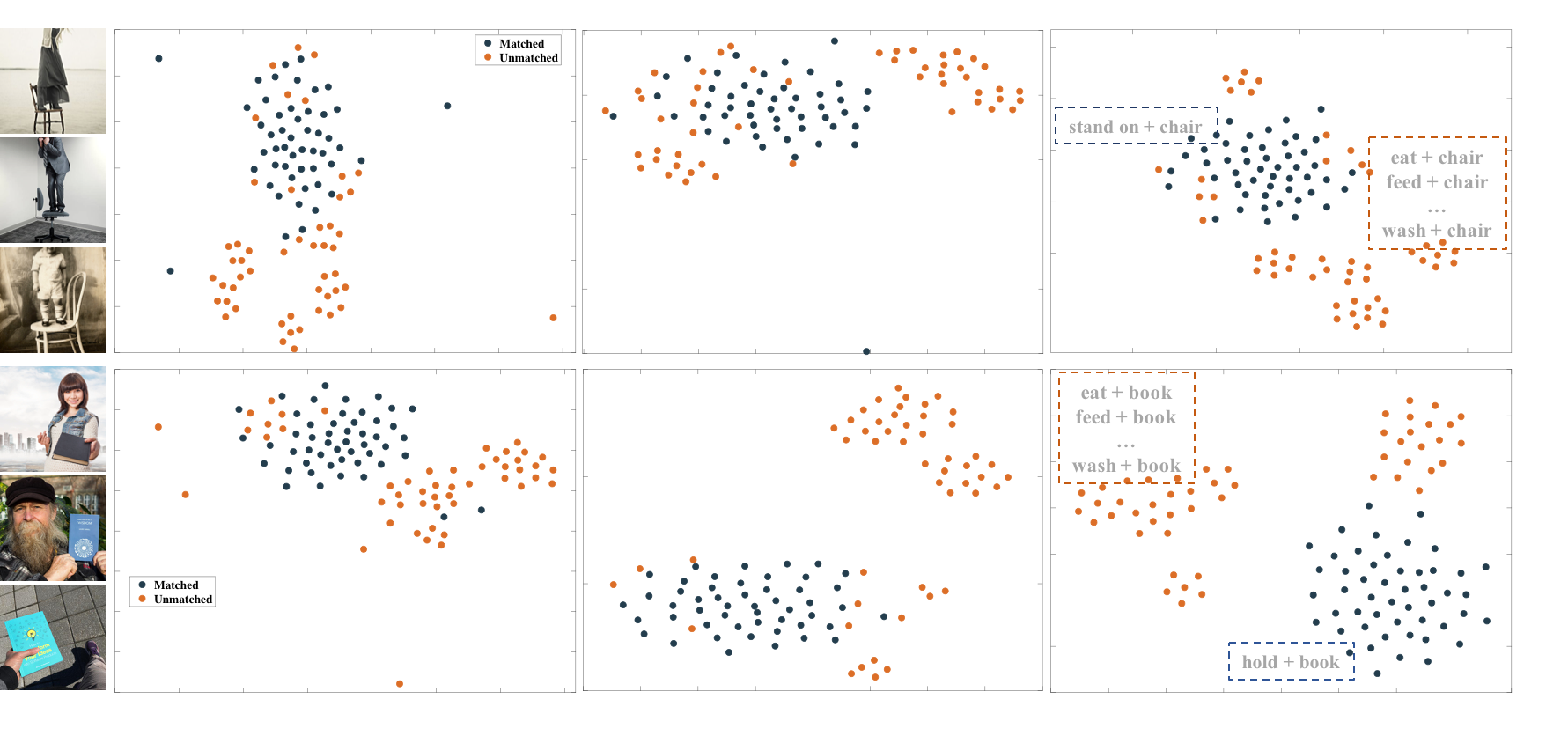}
\caption{{\bf $\bm{f_{V.Net}\oplus f_{Q.Net}}$ Feature Distribution Visualization (t-SNE): Matched vs. Unmatched.} Each colored sample represents a concatenated feature generated by {\it RGB} (\protect\cite{sung2018learning}) (left), {\it Orthovec2S} (center), or {\it S2S} (right). A ``Matched'' sample is generated by concatenating ($f_{V.Net}$) with corresponding ($f_{Q.Net}$). An ``Unmatched'' sample is generated by concatenating a $f_{V.Net}$ with one of the non-matching $f_{Q.Net}$, for instance, concatenating $f_{V.Net}$ of `hold + book' image with $f_{Q.Net}$ of `eat + book'.}
\label{fig:VNetQNet_MatchedUnmatched}
\end{center}
\end{figure*}

\noindent{\bf Q-Net + V-Net: Matched vs. Unmatched.} 
We considered how well the `Matched' and `Unmatched' (in terms of ground truth) features are segregated. As Figure \ref{fig:VNetQNet_MatchedUnmatched} suggests that S2S-driven $\bm{f_{V.Net}\oplus f_{Q.Net}}$ features provide better discriminative power which coaligns with previously shown quantitative comparison.

\section{Ablation Studies}
\label{sec:AblationStudies}

\subsection{Query generation for Q-Net}
\label{subsec:QNetGeneration}
We have tested several ways (element-wise sum, vertical concatenation, horizontal concatenation, Hadamard product) on how a single query vector is constructed from a pair of word representation vectors (V and O vectors) in {\it Q-Embed}. Table \ref{tab:queryComparison} shows that concatenating the two vectors along the vertical direction ({\it catV}) provides the best result in terms of zero-shot accuracy (yet, comparable to {\it sum}), suggesting that having semantic information separately encoded for V and O provides richer information and thus more effective.

\subsection{Separate encoding of V and O}
\label{subsec:SeparateVandO}
Instead of feeding a combined vector ($v_{V+O}$) into a single {\it Q-Net}, we implemented two networks for {\it Q-Net} to separately encode $v_{V}$ and $v_{O}$. As can be seen in Table \ref{tab:separateModules}, disintegrating V and O information is found to be more effective in most cases than having a single module for the query.


\begin{table}[!htb]
\small
\caption{{\bf Encoding verb and object query with separate modules.} RGB: Sung et al. \cite{sung2018learning}, OrthoVec2S: Orthonormal Vectors-to-space, S2S: Semantics-to-space. All the numbers indicate recognition accuracy in [\%]. Numbers in parentheses indicate single module-based.} 
\begin{center}
\begin{tabular}{cccc}
\toprule
Architecture & RGB & OrthoVec2S & S2S \\
\midrule
ResNet18 &  35.60 (33.53) & 45.47 (40.40) & {\bf 49.73} (46.27) \\
ResNet34 & 37.07 (38.00) & 46.93 (43.53) & {\bf 50.07} (48.87) \\
ResNet50 & 40.46 (41.73) & 43.87 (44.60) & 49.27 {\bf(50.47)} \\
\bottomrule
\end{tabular}
\end{center}
\label{tab:separateModules}
\end{table}

\begin{table}[!htb]
\small
\caption{{\bf Q-Net textual query generation comparison.} sum: word embeddings are summed, catV: word embeddings are concatenated vertically, catH: word embeddings are concatenated horizontally, Hadamard: word embeddings are multiplied in element-wise manner followed by L2 norm. ResNet50 is consistently used for the I-Nets in producing the accuracies shown in this table. All the numbers indicate recognition accuracy in [\%].} 
\begin{center}
\begin{tabular*}{\linewidth}{c@{\extracolsep{\fill}}ccc}
\toprule
sum & catV & catH & Hadamard \\
\midrule
50.47 & {\bf 52.13} & 40.47 & 24.53\\
\bottomrule
\end{tabular*}
\end{center}
\label{tab:queryComparison}
\end{table}

\section{Conclusion}
\label{sec:Conclusion}
We introduced a simple, yet powerful semantics embedding approach called Semantics-to-Space (S2S) designed to inference human-object-interaction images with verb-object query within two-stream ZSL architectures. Unlike the existing ZSL approaches, we augment the visual information by embedding the semantics in a spatial sense, thus providing more effectiveness in learning the matching VO queries and visuals in zero-shot scenarios. We also show that S2S can be used as a general module to enhance the performances of various ZSL baseline architectures.

\bibliographystyle{aaai}
\bibliography{mybib}

\end{document}